\documentclass{article}
\usepackage{spconf,amsmath,graphicx}
\usepackage{hyperref}
\usepackage{algpseudocode}
\usepackage{url}
\usepackage[ruled,vlined]{algorithm2e}
\usepackage{tabularx}
\usepackage{multirow}
\usepackage{makecell}
\usepackage{microtype}
\usepackage{colortbl}
\usepackage[colorinlistoftodos,textwidth=20mm]{todonotes}
\usepackage{mathtools}
\usepackage{booktabs}
\usepackage{enumitem}
\usepackage{caption}
\usepackage{breqn}
\usepackage{subcaption}
\usepackage{bbold}
\usepackage{stmaryrd}
\usepackage{xcolor}

\usepackage{soul}
\usepackage{wrapfig,lipsum,booktabs}
\usepackage{enumitem}

\usepackage{cite}

\newcommand{\jobid}[1]{\ignorespaces}

\definecolor{grey}{cmyk}{0.1,0.1,0.1,0.1}
\definecolor{orange}{cmyk}{0.1,0.7,0.5,0.2}

\title{Avoid Overthinking in Self-Supervised Models for Speech Recognition}

  \name{Dan Berrebbi, Brian Yan, Shinji Watanabe} 
 \address{Carnegie Mellon University}
\begin{document}
\ninept
\maketitle
\begin{abstract}
Self-supervised learning (SSL) models reshaped our approach to speech, language and vision. 
However their huge size and the opaque relations between their layers and tasks result in slow inference and \textit{network overthinking}, where predictions made from the last layer of large models is worse than those made from intermediate layers.
Early exit (EE) strategies can solve both issues by dynamically reducing computations at inference time for certain samples.
Although popular for classification tasks in vision and language, EE has seen less use for sequence-to-sequence speech recognition (ASR) tasks where outputs from early layers are often degenerate. 
This challenge is further compounded when speech SSL models are applied on out-of-distribution (OOD) data.
This paper first shows that SSL models do overthinking in ASR.
We then motivate further research in EE by computing an optimal bound for performance versus speed trade-offs.
To approach this bound we propose two new strategies for ASR: (1) we adapt the recently proposed patience strategy to ASR; and (2) we design a new EE strategy specific to ASR that performs better than all strategies previously introduced.

\end{abstract}
\section{Introduction}
Even though SSL models significantly improved state of the art performances on most speech tasks \cite{hubert, wav2vec2, wavLM, decoar, superb}, major drawbacks weaken their impact. 
First their huge stack of transformer layers \cite{attention_transformer} make \textbf{inference time quite slow}, which is non desirable for on-device problems for instance. 
Second as their training is task agnostic \cite{dery2022should}, each layer role and performance for ASR task remains unclear \cite{Pasad2021LayerWiseAO}. 
In particular such large models are prone to \textbf{overthinking} \cite{overthinking} which occurs when an accurate prediction is reached at an intermediate layer $i$ and computations of layers $j>i$ are wasteful and potentially destructive in terms of prediction quality.

\noindent \par  Orthogonal to static model compression strategies such as knowledge distillation \cite{NIPS2014_ea8fcd92,hinton2015distilling} or weight pruning \cite{janowsky1989pruning}, early exit methods \cite{branchynet} address the overthinking problem by \textbf{dynamically reducing computations} for certain samples at \textbf{inference time}. 
Typical approaches add lightweight exit heads on top of some layers. Those heads, called branches, compute exit scores usually based on confidence metrics such as entropy of softmax prediction distribution. 
During inference if an intermediate layer's exit head outputs an exit score bigger than a fixed threshold, the model outputs this layer's prediction and stops forward computation, saving precious time.

\noindent \par Despite their huge success in vision \cite{deecap} and language \cite{bert_looses_patience,xin-etal-2020-deebert,xin-etal-2021-berxit,liao-etal-2021-global,elbert, Early_Exiting_with_Ensemble_Internal_Classifiers} and their portfolio of potential applications, EE have not yet met a huge enthusiasm in speech \cite{Chen2021DontSB,Li2021LearningTI,9746863} and in particular in ASR \cite{hubert-EE}.
Early exiting is very challenging for ASR. 
First ASR is a complex sequence-to-sequence task in which models fail to get $100\%$ accuracy for most samples. On the contrary most EE approaches were designed for classification problems \cite{branchynet, bert_looses_patience} in which early predictions are very likely to be correct and constant across layers. 
This enabled \cite{bert_looses_patience}, inspired by early stopping strategies \cite{earlystopping}, to introduce a patience criteria which outperformed confidence based EE techniques on BERT \cite{devlin-etal-2019-bert}. 
Secondly, speech SSL models are commonly fine-tuned or used out of the box in \textbf{OOD setups} in which audio differs in language, noise level, type of speech and accent from the pre-training (and fine-tuning).
This frequent scenario in ASR research remains very challenging \cite{Hsu2021RobustW2} and it is legit to wonder whether EE strategies would be effective in such setup. 


The contributions of this work are summarized as follows:
\begin{itemize}[leftmargin=*,noitemsep]
    \item We show that ASR models do overthinking and analyze such phenomenon for in-domain and OOD scenario;
    \item We compute theoretical lower bounds of speed/quality trade-offs for EE strategies through dynamic programming;
    \item We adapt patience based EE strategies to suit ASR task;
    \item We introduce \textit{overlang}: a vocabulary based new EE strategy designed from our findings on overthinking in ASR.
\end{itemize}

\section{Motivations : Overthinking in ASR}

For an audio sample $x$ and a model composed of $N$ layers, we will note $\hat{y}_i(x)$ the output prediction of layer $i\in \{i_{\text{min}}, ..., N\}$. 
We assume that the model has exit heads on top of each layer except the $i_{\text{min}}$ first layers which predictions are often degenerated.
$\hat{y}_i(x)$ is a character sequence that can be compared to the reference transcription $y(x)$\footnote{For not overloading formulas we may write $\hat{y}_i$ and $y$.}. 
Following \cite{overthinking} a model overthinks for an input $x$ if
\begin{equation}
    \exists \; i < N \text{ such that } \text{error}(\hat{y}_i(x))\leq \text{error}(\hat{y}_N(x))
    \label{eq:overthinking}
\end{equation}
\noindent If Equation~\ref{eq:overthinking} is an equality then only time is wasted but performance is not impacted (we will call this scenario overthinking without degradation), but it could also be that quality of output is degraded.
EE strategies aim to find trade-offs between saving computations, so outputting $\hat{y}_i$ with a \underline{small $i$}, and having \underline{$\hat{y}_i$ quality} better than $\hat{y}_N$ (overthinking scenario) or at least not too much degraded.

\subsection{Experimental details}
 \label{sec:exp_details}

\noindent \textbf{Model -} 
We used Hubert Large model \cite{hubert} pre-trained (PT) on Librilight \cite{librilight} 60k hours and fine-tuned (FT) on Librispeech \cite{panayotov2015librispeech} 960 hours. 
We used English characters as token set. 
We added exit heads over all layers from layer 10. Our exit heads are made of a feed-forward module and a CTC \cite{CTC} head\footnote{Thus significantly smaller than \cite{hubert-EE} thus increasing computation savings.}.
Heads training (HT) was done on Librispeech 100h clean split, using the intermediate CTC loss \cite{intermediate_CTC}.
More details and hyperparmeters are released in \href{https://github.com/DanBerrebbi/espnet/tree/Hubert_EE}{ESPnet} \cite{watanabe2018espnet}.

\noindent \textbf{Data -} We used three datasets for inference: dev-clean and dev-other sets from Librispeech \cite{panayotov2015librispeech}, and the English dev set from Commonvoice 5.1\cite{commonvoice}\footnote{We call the sets respecitively \textit{Libri\_dev\_clean}, \textit{Libri\_dev\_other} and \textit{CV\_en}.}.
Table~\ref{tab:OOD} exhibits the distribution shifts for those sets, as accents, noise level and distributions (audiobooks/human generated sentences) differ.
We see that \textit{Libri\_dev\_clean} is fully in-distribution while \textit{Libri\_dev\_other} is partly OOD and \textit{CV\_en} is totally OOD.

\begin{table}[h!]
    \caption{Distribution shifts for the inference sets.} 
   
    \label{tab:OOD}
    \centering
\resizebox{0.9\linewidth}{!}{
    \begin{tabular}{c|ccc}
        \toprule
         & \textit{Libri\_dev\_clean} & \textit{Libri\_dev\_other} & \textit{CV\_en} \\
        \midrule
         PT \& FT & \textcolor{teal}{in distribution} & \textcolor{teal}{in distribution} & \textcolor{red}{OOD} \\
          HT& \textcolor{teal}{in distribution} &  \textcolor{red}{OOD} & \textcolor{red}{OOD} \\
        
        \bottomrule
    \end{tabular}
}
\end{table}

\vspace{-1em}

\subsection{Do speech SSL model overthink? \label{ssec:overthink}}

Large networks such as SSL models are very prone to overthinking as shown in \cite{overthinking,bert_looses_patience,deecap}. 
However, those studies mostly targeted classifications tasks such as sentiment analysis \cite{bert_looses_patience}.
For such task, it seems intuitive that for some $i$, layers $j\geq i$ will output the same class \cite{bert_looses_patience}. 
On the other hand ASR is a complex sequence to sequence task in which the model's layers are refining a predicted sentence. 
In most cases $\hat{y}_i(x)$ is different from $y(x)$ for all $i\in \llbracket 1,N\rrbracket$. Thus we can think that overthinking does not occur since refining the prediction until the last layer could be more beneficial than harmful due to the complexity of the task. 


\vspace{-1em}
\begin{figure}[h!]
    \centering
    \includegraphics[scale=0.5]{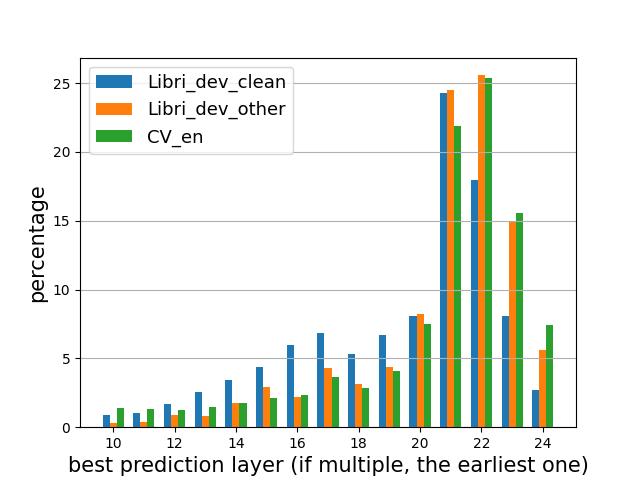}
    \caption{Overthinking in ASR for in-domain and OOD sets.}
    \label{fig:overthinking}
\end{figure}
  


\noindent In Figure~\ref{fig:overthinking} we show that SSL models do overthinking for ASR: bar $i$ is the percentage of samples per set for which the best prediction (for word error rate, WER) is reached \textbf{first} at layer $i$.
We remark that :

\begin{itemize}[leftmargin=*,noitemsep]
    \item For all sets less than $10\%$ of the samples need to go to the last layer to reach best prediction, which is a clear case of overthinking.
    \item For \textit{Libri\_dev\_clean} a non negligible portion of the samples reach their best prediction very early (i.e. before layer $20$) whereas for the two OOD sets (\textit{Libri\_dev\_other} and \textit{CV\_en}) more than $70\%$ of the samples need to wait the last 4 layers to reach best prediction.
\end{itemize}

\noindent Additionally to Figure~\ref{fig:overthinking}, our aim is to differentiate if the model is overthinking without degradation (so only wasting time) or is wasting time \textbf{and} degrading performances. 
In other terms, among samples that reach best prediction before the last layer what is the share for which the last layer remains the best one\footnote{This cannot be infered from Figure~\ref{fig:overthinking}, we conducted additional analysis.}?
We found that it is more than $90\%$ for the Librispeech sets (so mainly overthinking without degradation) whereas for \textit{CV\_en} more than $22\%$ of the samples predictions are degraded when reaching the last layer.



\subsection{Lower bound for early exit strategies \label{ssec:bounds}}

\begin{figure}[h!]
      \centering
      \includegraphics[scale=0.25]{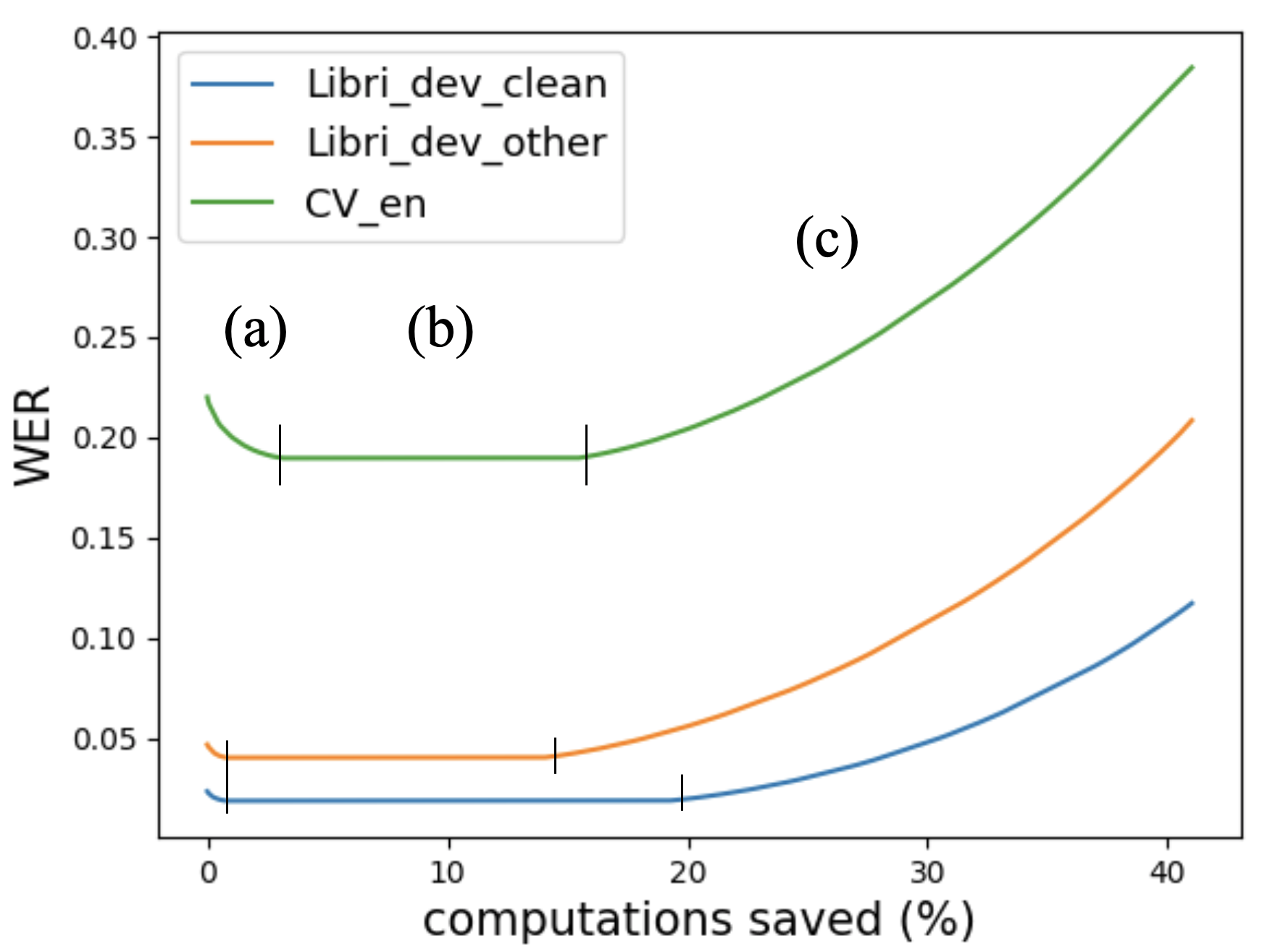}
      \caption{Theoretical bounds to early exits trade-offs}
      \label{fig:theoretical_bound}
  \end{figure}

Using every layer predictions for each sample we computed the best performances for any reachable number of saved computations through dynamic programming. 
Optimum are \textbf{convex} (see Figure~\ref{fig:theoretical_bound}) for the 3 datasets showing that (up to some point) computation savings increase relatively faster than quality degradation which is desirable.



\noindent We remark that all curves are composed of \textbf{(a)} a small decreasing portion followed by \textbf{(b)} a non negligible plateau and finally \textbf{(c)} an increasing part. Portion (a) corresponds to a part where we can save computing time \textbf{and} increase quality of outputs. This is the evidence of quality degradation due to overthinking. 
As mentioned in Section~\ref{ssec:bounds} \textit{CV\_en} suffers more from this phenomenon and so phase (a) is longer for this set in Figure~\ref{fig:theoretical_bound}. 
The plateau portion (b) corresponds to overthinking without degradation : wasteful but not harmful computations.
Figure~\ref{fig:theoretical_bound} optimum reflects Figure~\ref{fig:overthinking} findings : the in-distribution set (\textit{Libri\_dev\_clean}) is more prone to overthinking so this plateau is longer for this set. For this set, an optimal strategy can save up to $20\%$ of computations with no degradation (on average).
Finally (c) corresponds to the trade-off phases where no more computations can be saved with no degradation cost. 
In that part we see that the slope is slightly lower for the in-domain set so that computations will be cheaper to save for \textit{Libri\_dev\_clean} than for the OOD sets.

\textbf{Note on computation saved - } In network compression studies speed is often calculated through wall clock time or floating point operations saved. 
However as computations for transformer blocks are quadratic in the input length while the total number of errors is more likely to be linear in the length of the sentence we argue that reporting those time measurements is not desirable for EE.
Indeed a strategy exiting long utterances very early and small ones late would be better than the reverse strategy because it would suffer a linear degradation of performances but would get a quadratic gain of time.
As we do not want the strategy to favor short utterances, we rather compute relative gain of time for each sample that is simply reported as the proportion of layers that were skipped\footnote{As this could have the opposite effect, i.e. favor very early exit for short sentences, we will report results for sentences of more than $10$ words only.}.

\section{Proposed methods}
\label{sec:proposed_methods}

We formally note $\zeta$ the criterion (taking only boolean values) such that for a sample $x$ the model exits at layer $i^*=\min_{i}\{i | \zeta(\hat{y}_i(x))=1\}$.
We note $T(x)$ (simplified as $T$) the number of frames of sample $x$ and $C$ the number of tokens. 
$f_{i,t,c}(x)$ is the output of the softmax of exit branch at layer $i$, for frame $t\in \llbracket1,T\rrbracket$ and token $c\in \llbracket1,C\rrbracket$\footnote{Our models are CTC-based only however the strategies can be applied to encoder-decoder or any other architecture.}. \\



\subsection{Confidence based methods \label{sec:confidence_based_methods}}

We first tried the commonly used entropy based confidence score \cite{branchynet}, computed from the softmax output following Eq.~\ref{eq:entropy_confidence}. 
\begin{equation}
    \text{score($\hat{y}_i(x)$)}= - \frac{1}{T\cdot C} \sum_{t=1}^{T} \sum_{c=1}^{C} f_{i,t,c}(x) \cdot \log(f_{i,t,c}(x))
    \label{eq:entropy_confidence}
\end{equation}

\noindent Following \cite{hubert-EE} we also tried a variant where the confidence score is given by the average over frames of the maximum softmax(ed) probability instead of the entropy of the distribution. As a well calibrated model would be confident about accurate prediction, our criteria is is given by Equation~\ref{eq:entrop_confidence_thresh}, 

\begin{equation}
    \zeta(\hat{y}_i(x)) = \mathbb{1}_{\text{score}(\hat{y}_i(x))<\tau} 
    \label{eq:entrop_confidence_thresh}
\end{equation}

\noindent where $\tau$ is a fixed threshold and $\mathbb{1}$ is the indicator function\footnote{For the maximum probability variant, the inequality sign is changed.}.







\subsection{Patience based methods \label{sec:patience_based_methods}}

Strategies introduced in Section~\ref{sec:confidence_based_methods} can however be weak when a model is overconfident or poorly calibrated and cannot handle regression cases \cite{bert_looses_patience}. 
Inspired by early stopping, PABEE \cite{bert_looses_patience} solved those issues by introducing patience based strategies. 
The principle is that criteria $\zeta$ allows exit at layer $i$ if the prediction at layer $i$ is consistent with the predictions of the few layers before $i$. 
It enables to (1) avoid unstable predictions; (2) exit when the model somehow converged to a prediction; and (3) indirectly predict from an ensemble of layers. 
Formally with a distance $d$, a threshold $\tau$ and a patience threshold $\rho\in \mathbb{N}$, the patience strategy can be written as in Equation~\ref{eq:patience_definition}. 
\begin{equation}
    \zeta(\hat{y}_i(x)) = 1 \Leftrightarrow \forall j\in \llbracket i-\rho,i \rrbracket , \; d(\hat{y}_j(x), \hat{y}_{j-1}(x)) < \tau 
    \label{eq:patience_definition}
\end{equation}

\noindent Distances proposed in \cite{bert_looses_patience} (e.g. L2 distance of two predictions) cannot be applied to sequential outputs. 
We used two distances for patience strategies : \textbf{cross-entropy} at the logits level and \textbf{Levenshtein distance} at the output sentence level.

\subsection{Overlang : a vocabulary based strategy \label{sec:language_based_methods}}


Finally, we propose a novel criterion that directly targets the EE aim:  \textbf{output high quality predictions} while \textbf{avoiding overthinking}.
As we stick to Hubert configuration \cite{hubert} we adopt characters as our token set.
Predicted words (sequences of letters delimited by spaces) may thus be any combination of characters which are potentially out-of-vocabulary (e.g. due to misspelling)\footnote{We note that this findings and method also apply for any type of subword.}.
A first observation is that when an early layer prediction is of poor quality, most of its \textbf{incorrect words are out-of-vocabulary}. 
Table~\ref{tab:explain_overlang} provides examples of predicted sentences from early layers with several of such out-of-vocabulary errors. 
These predictions are phonetically proximate to the reference, but consist of many invalid words -- we can reasonably detect these by checking against a known vocabulary.

\begin{table}[h!]
    \caption{
    Example ASR errors flagged by an out-of-vocabulary check. 
    } 
   
    \label{tab:explain_overlang}
\resizebox{\linewidth}{!}{
    \begin{tabular}{|l|l|}
        \hline
         Reference Sentence &  Predicted sentence \\
        \hline
        \hline
        now active exploitation was required &  now \textcolor{red}{actiev} \textcolor{red}{expoitation} was \textcolor{red}{requie} \\
        \hline
        he asked on seeing the prisoners &  he \textcolor{red}{ased} on seeing the \textcolor{red}{prisiners} \\
        
        \hline
    \end{tabular}
}
\end{table}

 Overthinking happens only when accurate predicted words (which are in-vocabulary) are modified by later layers and so cannot occur on words that are out-of-vocabulary (under assumption that out-of-vocabulary words are incorrect)\footnote{This assumption is acceptable given that proportion of out-of-vocabulary words in reference transcription (e.g. named entities) is quite small.}. 
As our model was never guided by any language model in its training, we assume that when early layers predict a word that is in vocabulary this word is very likely to be the accurate prediction. 
Overthinking may then happen if this correct word is changed to another in or out-of-vocabulary word.

This leads us to the following criteria : exit is allowed at layer $i$ if the prediction of this layer does not contain too many out of vocabulary words. 
We note $\mathcal{V}$ our English vocabulary\footnote{Given by any English dictionary, we used \href{https://pypi.org/project/english-words/}{english-words librairy}.}.
We define $W(\hat{y}_i(x))$ in Equation~\ref{eq:card_non_english_words} as the proportion of in-vocabulary words in layer $i$'s prediction, where $\text{card}(.)$ is the cardinal of a set and \text{length}(.) is the number of words of a sentence.
\vspace{-.5em}
\begin{equation}
    W(\hat{y}_i(x)) = \frac{\text{card}(\{w\in \hat{y}_i(x) \text{ such that } w\in \mathcal{V}\})}{\text{length}(\hat{y}_i(x))}
    \label{eq:card_non_english_words}
\end{equation}

\noindent For a threshold $\tau$, an exit criterion can be derived as in Equation~\ref{eq:criteria_non_english_words}:
\begin{equation}
    \zeta(\hat{y}_i(x)) = \mathbb{1}_{W(\hat{y}_i(x))\geq \tau} 
    \label{eq:criteria_non_english_words}
\end{equation}

\noindent 

\noindent This criterion is orthogonal to the ones defined in Section~\ref{sec:confidence_based_methods} and \ref{sec:patience_based_methods} so that they also can be combined together.
We combined $\zeta$ defined in Equation~\ref{eq:criteria_non_english_words} with a patience criterion on the $W(\hat{y}_i(x))$ values.
For a chosen integer $\rho$, if $W(\hat{y}_j(x))$ is constant for $j\in \llbracket i-\rho , i\rrbracket$ then the model exits at layer $i$ even if  $W(\hat{y}_j(x))<\tau$. 
For a given $\rho$ and $\tau$, Equation~\ref{eq:overlang} defines \textit{overlang}, our vocabulary based EE criteria, 
\begin{equation}
    \zeta(\hat{y}_i) = \text{max}(\mathbb{1}_{W(\hat{y}_i)\geq \tau} \;,\; \mathbb{1}_{W(\hat{y}_{i})=W(\hat{y}_{i-1})=...=W(\hat{y}_{i-\rho})})
    \label{eq:overlang}
\end{equation}

\noindent Where max is equal to one if at least one of the indicators is satisfied.

\section{Results and Analysis}

\subsection{Additional details on experiments}

Table~\ref{tab:hyperparams} gathers the values we used for thresholds in our EE strategies. 

\begin{table}[h!]
    \caption{Experimental details on EE strategies} 
   \centering
    \label{tab:hyperparams}
\resizebox{0.95\linewidth}{!}{
    \begin{tabular}{|lll|}
        \hline
         Strategy & Measure  & range of hyperparameters \\
        \hline
        \hline
         Confidence & Entropy  & $\tau \in \{0.002, 0.0025, 0.003, ... 0.006 \}$ \\
         \hline
         Confidence & Maximum Probability  & $\tau \in \{0.93, 0.935, 0.94, ... 0.97 \}$ \\
         \hline
         \multirow{2}{*}{Patience} & \multirow{2}{*}{Cross-Entropy} & $\tau \in \{0.1, 0.2, 0.3, 0.4, 0.5 \}$ \\
          & & $\rho \in \{1, 2, 3, 4, 5\}$ \\
          \hline
        \multirow{2}{*}{Patience} & \multirow{2}{*}{Levenshtein distance} & $\tau \in \{0.05, 0.07, 0.1, 0.15, 0.2 \}$ \\
          & & $\rho \in \{1, 2, 3, 4, 5\}$ \\
          \hline
        \multirow{2}{*}{Overlang} & \multirow{2}{*}{-} & $\tau \in \{0.6, 0.625, 0.65 ... 0.95 \}$ \\
          & & $\rho=2$ \\
        
        \hline
    \end{tabular}
}
\end{table}

\subsection{Efficiency of the strategies}

\begin{figure}[h!]
    \centering
    \includegraphics[scale=0.52]{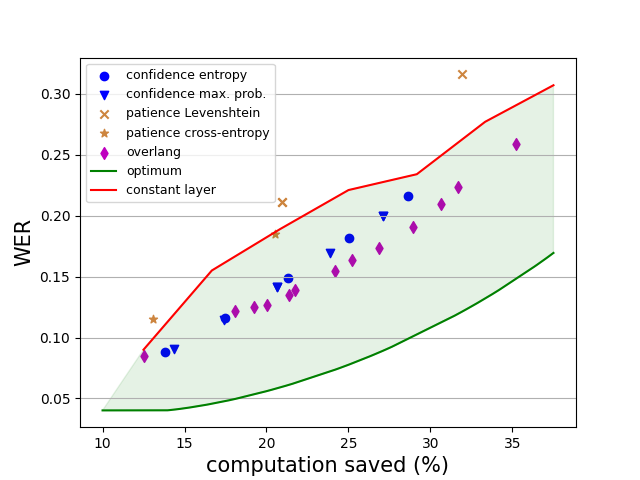}
    \caption{Trade-off comparison for different EE strategies}
    \label{fig:dev_other_final_curve}
\end{figure}

Figure~\ref{fig:dev_other_final_curve} presents the speed/performance trade-offs of the EE strategies introduced in Section~\ref{sec:proposed_methods} for \textit{Libri\_dev\_other} set\footnote{Similar behaviors are observed on the two other inference sets.}. 
We add in green the optimal bound curve computed in Section~\ref{ssec:bounds} and in red the curve linking the trade-offs reached by the naive strategies: \textit{always exit at a fixed layer $i$} (i.e. $\zeta_i(\hat{y}_j(x)) = \mathbb{1}_{j=i}$). 
Those $\zeta_i$ are naive strategies as they exit at the same layer for every input sample. 
Thus we use them as an informal upper bound: any EE strategy should reach better trade-offs than those naive methods. 
We color in green an 'acceptable trade-offs zone' which is in between the optimal bound and the naive trade-offs. 
Under $10\%$ of computations saved, the gap between optimal bound, proposed methods and constant layer strategies is very small and this range of time savings is of limited interest for potential applications.
On the other hand, performances with large time savings are too much degraded to be applicable in the real world.
We thus plot the $10\% \text{ to } 40\%$ computations saved zone.

\noindent We first remark that the patience based strategies (both with cross-entropy and Levenshtein distance) give bad trade-offs compared to confidence, \textit{overlang}, and even naive strategies. 
Confidence and \textit{overlang} strategies give better results than the naive strategies and \textit{overlang} outperforms the trade-offs obtained by confidence-based strategies quite consistently over the considered zone. 
Table~\ref{tab:trade_offs_comparison} presents a comparison of Word Error Rate (WER) obtained by confidence and \textit{overlang} for a same (or very similar) amount of computations saved (vertically aligned points on Figure~\ref{fig:dev_other_final_curve}).
We remark that \textit{overlang} indeed reaches better trade-offs than confidence based approaches for this range of savings with WER reduction \footnote{WER reduction is given by : $100\cdot \frac{\text{WER(confidence)-WER(overlang)}}{\text{WER(confidence)}}$} of about $10\%$.

\begin{table}[h!]
    \caption{Trade-offs comparisons for confidence-based strategies and \textit{overlang}. We also compute relative reductions of the WER.}
   
    \label{tab:trade_offs_comparison}
\resizebox{\linewidth}{!}{
    \begin{tabular}{l|ccccc}
        \toprule
         & \multicolumn{5}{c}{\underline{\textsc{Computation Saved}}} \\
         Method & $\sim 21.3\%$ & $\sim 24.0\%$ & $\sim 25.1\%$ & $\sim 27.0\%$ & $\sim 28.7\%$   \\
        \midrule
        Confidence (WER)  & $14.9$ & $16.9$ & $18.2$ & $20.0$ & $21.6$ \\

        \textit{Overlang} (WER)  & $13.5$ & $15.5$ & $16.4$ & $17.4$ & $19.0$ \\
        \midrule
        \textbf{Relative Reduction} & $\textbf{9.4\%}$ & $\textbf{8.3\%}$ & $\textbf{9.9\%}$ & $\textbf{13.0\%}$ & $\textbf{12.0\%}$ \\

        \bottomrule
    \end{tabular}
}
\end{table}

\subsection{Overthinking for Patience, Confidence and Overlang}

We confirm intuitions from Figure~\ref{fig:dev_other_final_curve} by computing overthinking percentage for three strategies saving approximately $21\%$ computations\footnote{cross-entropy patience with $\tau = 0.05 \text{ and } \rho=3$, entropy confidence with $\tau=0.055$ and \textit{overlang} with $\tau=0.8$.}. 
For a sample $x$ we note $l(x)$ the exit layer chosen by the strategy. 
We report overthinking by computing the proportion of $x$ for which $\exists \; i<l(x)$ such that $\text{WER}(\hat{y}_i(x))\leq \text{WER}(\hat{y}_{l(x)}(x))$.
Overthinking happens with the patience strategy at $68\%$ whereas confidence based strategy only gets $53\%$ and \textit{overlang} $49\%$.

 Utterance \textit{1630-96099-0016} from \textit{Libri\_dev\_other} provides an example of how the strategies can behave (see Table~\ref{tab:analyse}).

\begin{table}[h!]
    \caption{Early exit layers for utterance \textit{1630-96099-0016}.} 
   
    \label{tab:analyse}
    \centering
\resizebox{0.95\linewidth}{!}{
    \begin{tabular}{|lcl|}
        \hline
         Strategy & Exit Layer  & Sentence \\
        \hline
        \hline
        Reference & - & \textit{he left everything behind}\\
        \hline
         Patience (cross-entropy) & 24 & \textit{he left everything behind}\\
         \hline
         Confidence (entropy) & 13  & \textit{he left \textcolor{red}{everthing} behind} \\
         \hline
        Overlang & 15  & \textit{he left everything behind} \\
        
        \hline
    \end{tabular}
}
\end{table}

\noindent First we see that the language strategy benefits over the confidence one as \textit{everthing} is not in-vocabulary: \textit{overlang} exits two layers later only but with a correct sentence.  
On the contrary, the patience strategy using cross-entropy exits at the last layer as the output probability distribution is not stable (in terms of entropy), thus overthinking a lot. 

\subsection{Additional remarks on experiments}

We add empirical findings of practical interest for the readers:
\begin{itemize}[leftmargin=*,noitemsep]
    \item Combining patience and confidence strategies does not improve over confidence ones contrary to combinations of patience and \textit{overlang}.
    A plausible explanation is that when prediction is constant across layers, \textit{overlang} ratio $W$ of in-vocabulary words is constant and so if $W<\tau$  the model do not exit early.
    On the other hand, confidence based methods are not impacted by this scenario because even if predicting the same sentence across different layers, the model usually becomes more confident about its prediction and thus quickly reaches the confidence threshold and exits.    
    \item No behavioral difference was found between in-distribution and OOD. It demonstrates the feasibility of EE strategies for OOD.
\end{itemize}

\section{Conclusion and Future Directions}

We demonstrate that overthinking happens in ASR for both in-domain and OOD scenario. 
Our introduced \textit{overlang} strategy as well as confidence-based methods enable reaching good speed/quality trade-offs thus avoiding overthinking for some samples. 
We believe that research in this topic is very promising for a wide range of applications such as on-device systems or semi-supervised ASR where pseudo-labels could be generated (more rapidly) by early layers for some samples as \cite{berrebbi2022continuous} showed that bootstrapping from poor quality pseudo-labels is viable.

\section{Acknowledgments}
We would like to thank Navdeep Jaitly, Tatiana Likhomanenko and Ronan Collobert for helpful discussions on the topic. This work used the Extreme Science and Engineering Discovery Environment (XSEDE) ~\cite{xsede}, which is supported by National Science Foundation grant number ACI-1548562. Specifically, it used the Bridges system ~\cite{nystrom2015bridges}, which is supported by NSF award number ACI-1445606, at the Pittsburgh Supercomputing Center (PSC).

\clearpage

\bibliographystyle{IEEEbib}
\bibliography{mybib}

\end{document}